# LIGHTCAM: A FAST AND LIGHT IMPLEMENTATION OF CONTEXT-AWARE MASKING BASED D-TDNN FOR SPEAKER VERIFICATION

*Di CAO[1], Xianchen Wang[1], Junfeng Zhou[1], Jiakai Zhang[1], Yanjing Lei[1], Wenpeng Chen[2]*

[1]Zhejiang University of Technology, Hangzhou, China
[2]Hangzhou Jet-voice Technology Co., Ltd., Hangzhou, China

## ABSTRACT

Traditional Time Delay Neural Networks (TDNN) have achieved state-of-the-art performance at the cost of high computational complexity and slower inference speed, making them difficult to implement in an industrial environment. The Densely Connected Time Delay Neural Network (D-TDNN) with Context Aware Masking (CAM) module has proven to be an efficient structure to reduce complexity while maintaining system performance. In this paper, we propose a fast and lightweight model, LightCAM, which further adopts a depthwise separable convolution module (DSM) and uses multi-scale feature aggregation (MFA) for feature fusion at different levels. Extensive experiments are conducted on VoxCeleb dataset, the comparative results show that it has achieved an EER of 0.83 and MinDCF of 0.0891 in VoxCeleb1-O, which outperforms the other mainstream speaker verification methods. In addition, complexity analysis further demonstrates that the proposed architecture has lower computational cost and faster inference speed.

## 1. INTRODUCTION

A speaker verification system consists of two main components: an embedding extractor that transforms a random-length utterance into a fixed-dimensional speaker embedding, and a back-end model that computes the similarity score between the embeddings [1].

Speaker verification systems based on deep learning methods have made remarkable improvements over the past few years [2, 3, 4, 6, 7, 8]. One of the most popular systems is x-vector, which adopts time delay neural network (TDNN) as backbone. TDNN has a global sensor space over the feature channel through one-dimensional convolution operation. In order to capture details in the full frequency domain, usually a large number of filter coefficients are required. For example, ECAPA-TDNN [3] convolutional layer has a maximum of 1024 channels so as to achieve an optimal performance.

In addition, another improvement method for x-vector is to introduce two-dimensional convolutional neural network (CNN) and residual connections on ResNet [5]. These ResNet-based systems [6, 7] have achieved outstanding results. It captures local frequency and time domains information using two-dimensional convolution with small receptive fields. By reusing local feature patterns, precise details of frequency can be captured with fewer feature channels.

Both TDNN and ResNet perform well in speaker verification, and as the model depth increase, the network's performance continues to improve [4, 7]. However, these networks have high computational requirements and are not efficient in terms of inference speed. D-TDNN [9], an efficient speaker embedding model based on TDNN, is proposed as one of the solutions. It consists of Feedforward Neural Network (FNN) and TDNN layers, and adopts dense connection method. It has also achieved better results with fewer parameters.

Additionally, the attention mechanism has been widely applied in the field of speaker verification. Squeezing Excitation (SE) [10] technology compresses global spatial information into channel descriptors to model dependencies and recalibrate filter responses. At the same time, the time-pooling technique is further improved through the calculation of weighted statistics by the soft self-attention mechanism. ECAPA-TDNN unifies one-dimensional Res2Block with squeeze excitation and expands the temporal context of each layer to achieve a significant improvement. In [11] an attention-based context-aware masking (CAM) module is proposed to improve the performance of D-TDNN by focusing on interested speakers and blurring irrelevant noise. On this basis, a novel CAM++ network [12] is proposed to enhance CAM module and D-TDNN architecture. This CAM module uses multi-granularity pooling combined with global average pooling and segmented average pooling. It also increases the depth of the D-TDNN network and simultaneously reduces the channel size of each layer of filters to control the complexity of the network.

Although CAM++ has achieved good results in accuracy and efficiency, in practical speaker verification application scenarios, the computational complexity and inference time of the network are more important. In order to further enhance the advantages of CAM++ in implementation, we applied two improved methods to CAM++ and proposed LightCAM. Firstly, we use depthwise separable convolution module (DSM) instead of the original front-end convolution module (FCM). It aims to capture more detailed time-frequency details on acoustic feature maps with less computational complexity. Secondly, multi-scale feature aggregation (MFA) aggregates the outputs of each dense block in the channel dimension, which is beneficial for extracting multi-scale speaker features and improving the model's ability to discriminate between speakers.

This paper is organized as follows: Section 2 describes the proposed method, Section 3 presents experiments and results, and Section 4 summarizes and looks to future work.

## 2. PROPOSED METHOD

The LightCAM model consists of three parts: DSM for feature extraction, TDNN as backbone and MFA for feature fusion. Inspired by the depthwise separable convolution module used in MobileNet [13], the DSM is applied prior to the input of the backbone network, allowing more detailed time-frequency details to be captured on acoustic feature maps with less computational complexity. The output feature map is flattened along the channel and frequency dimensions. The D-TDNN backbone consists of three dense blocks of 12, 24 and 16 D-TDNN layers respectively, with CAM modules in each layer. In the end, MFA concatenates the outputs of each dense block in the channel dimension.

### 2.1. Depthwise Separable Convolution Module

DSM uses depthwise separable convolution to reduce model parameters and computational complexity, improving model efficiency in resource-constrained scenarios such as mobile devices. Unlike traditional convolution operations, depthwise separable convolution operation has two steps: depthwise convolution and pointwise convolution. Depthwise convolution has one convolution kernel responsible for each channel of the input data, and thereby produces a feature map with exactly the same number of channels as the input. The pointwise convolution operation is similar to the regular convolution operation in that it has a convolution kernel of size 1×1×M, where M is the number of channels in the previous layer. So here the convolution operation combines the maps from the previous step, weighted in channel direction, to create a new feature map. The number of parameters in depthwise separable convolution is about 1/3 of that of regular convolution, and it also has lower computational complexity than regular convolution.

Therefore, DSM uses a structure of multiple DSM-ResBlocks equipped with residual connection. The structure of individual DSM-ResBlock is shown in Figure 2. In DSM-ResBlock-1 and DSM-ResBlock-2, the output channel of depthwise separable convolution is set to 32, while in DSM-ResBlock-3 and DSM-ResBlock-4 the output channels are set to 64. The number of groups for depthwise convolution is the same as the number of input channels, while pointwise convolution is implemented on the output channels of depthwise convolution. Followed after each depthwise separable convolution, BatchNorm layer and ReLU activation function are applied. In DSM, the step size of three convolution operations in the frequency domain is set to 2, resulting in an 8-fold downsampling of acoustic features in the frequency domain dimension. The output feature map from the DSM is then

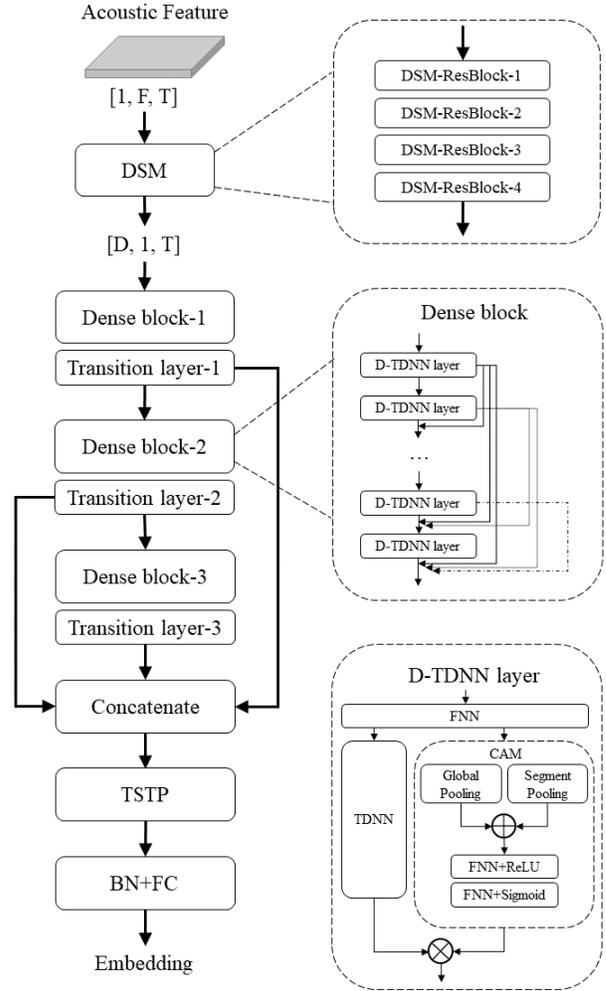

**Fig.1**. Proposed Structure of LightCAM. The model sequentially connects DSM, D-DTNN and MFA.

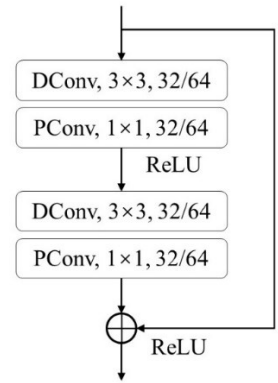

**Fig.2**. Proposed Structure of DSM-ResBlock. Each DSM-ResBlock consists of two sets of depthwise separable convolutions. (DConv denotes depthwise convolution and PConv denotes pointwise convolution).

flattened along the channel and frequency dimensions as input to the D-TDNN backbone.

## 2.2. D-TDNN with context-aware masking

The D-TDNN backbone consists of three dense blocks, with 12, 24, and 16 D-TDNN layers in each dense block. The growth rate of the D-TDNN layer in each dense block is set to 32 to ensure the complexity advantage of the network. In addition, each D-TDNN layer contains CAM modules, which serve as a simple and efficient pooling method to extend a single global pooling to multi granularity pooling, thereby aggregating contextual information from different levels. The CAM module can capture contextual information locally and globally with fewer parameters and high computational efficiency. It generates a mask through calculation to extract more effective information from the features output by TDNN. Specifically, the CAM module collects the hidden features output $X$ from the FNN, and then use global average pooling to extract global level contextual information $e_g$ while extracting segment level information $e_s$ via segment average pooling.

$$e_g = \frac{1}{T} \sum_{t=1}^{T} X_t \quad (1)$$

$$e_s^k = \frac{1}{s_{k+1} - s_k} \sum_{t=s_k}^{s_{k+1}-1} X_t \quad (2)$$

where $X_t$ denotes the $t$-th frame of $X$, $T$ is the length of $X$. And $s_k$ is the starting frame of the $k$-th segment of feature $X$. The frame level feature $X$ will be divided into continuous 100 frame segments with fixed length and segment average pooling will be applied to each segment. Therefore, context vector mappings at different levels ($e_g$ and $e_s$) are aggregated to predict context aware mask $M$.

$$M_t^k = \sigma(W_2 \delta(W_1(e_g + e_s^k) + b_1) + b_2),$$
$$s_k \leqslant t < s_{k+1} \quad (3)$$

where $\delta(\cdot)$ denotes the ReLU activation function, $\sigma(\cdot)$ denotes the Sigmoid activation function. Both $W_1$ and $W_2$ denote the weights of FNN. $b_1$ and $b_2$ denote the biases of FNN.

The resulting mask $M$ is used to mask the features by element multiplication, and a refined feature representation is generated.

## 2.3. Multi-scale Feature Aggregation

Recent research [14, 15] has shown that low-level feature mapping can also help to extract speaker vector maps accurately. The feature maps output by each dense block are concatenated and then fed as an input to the BatchNorm layer.

$$D = \text{Concat}(d_1, d_2, \ldots, d_L) \quad (4)$$

where $D$ represents the aggregated output features, $d_L$ is the output characteristics of the $L$-th level D-TDNN block.

Afterwards, temporal statistics pooling (TSTP) [2] is used to obtain key information for each frame, enhancing the representation ability of features and improving the accuracy of the speaker vector maps. The feature mapping is processed through BatchNorm layer (BN) and fully connected layer (FC) to obtain the final speaker embedding.

## 3. EXPERIMENTS AND RESULTS

### 3.1. Dataset

Experiments primarily rely on the well-established benchmark datasets Voxceleb1 and Voxceleb2 [15, 16, 17]. The audio data in these datasets is recorded at 16kHz sample rate, stored in 16-bit, mono, PCM-WAV format. We use the Voxceleb2 development set for training, which comprises 1,092,009 utterances from 5,994 speakers. Test set consists of three clean subsets from Voxceleb1, namely Voxceleb1-O, E, H. These three sets are widely recognized standards for evaluating the performance of speaker recognition models. Furthermore, data augmentation, the use of the RIR [18] dataset to simulate reverberation, and the addition of noise from the MUSAN [19] are all incorporated into training process.

### 3.2. Experiment setup

In the experiments, 80-dimensional F-bank features are utilized as the input, which are extracted with 25ms window every 10ms. In order to enhance the generalization ability of the system, speed perturbation augmentation is applied by randomly sampling a ratio from {0.9, 1.0, 1.1}. The processed audio is considered to be from a new speaker [21].

Same as mainstream speaker verification systems, angular additive margin softmax (AAM-Softmax) loss [22] is used for all experiments. The margin and scaling factors of AAM-Softmax loss are set to 0.2 and 32 respectively. Stochastic gradient descent (SGD) optimizer is adopted with an exponential decay scheduler and a linear warm-up scheduler, where the learning rate is varied between 0.1 and 5e-5. The momentum is 0.9, and the weight decay is 1e-4. The input audio during training is a small batch of samples randomly cropped for 3 seconds from the samples.

In the evaluation process of the system, ECAPA-TDNN [3], ResNet34 [6], and CAM++ [12] are used as benchmarks. We use cosine similarity scoring for evaluation, without applying any score normalization in the back-end, and large margin fine-tuning strategy [23] is not used either. For evaluation index, we use equal error rate (EER) and the minimum detection cost function (MinDCF) with $p_{Target} = 0.01$, $C_{FA} = C_{Miss} = 1$. In addition, the floating-point operations (FLOPs) and real-time factor (RTF) of models are also tested to analyze the complexity of system.

Table.1. Performance comparison of different network structures on VoxCeleb1 dataset

| Architecture | Params(M) | VoxCeleb1-O EER(%)/MinDCF | VoxCeleb1-E EER(%)/MinDCF | VoxCeleb1-H EER(%)/MinDCF |
|---|---|---|---|---|
| ECAPA-TDNN | 14.65 | 0.86/0.0921 | 1.07/0.1185 | 2.06/0.1956 |
| ResNet34 | 6.63 | 0.86/0.0912 | 1.05/0.1214 | 1.96/0.1921 |
| CAM++ | 7.18 | 0.88/0.0935 | 0.97/0.1183 | 1.89/0.1971 |
| **LightCAM** | **8.15** | **0.83/0.0891** | **0.95/0.1114** | **1.86/0.1922** |

Table.3. Ablation study of LightCAM modules. With DSM and MFA applied on CAM++, we get LightCAM.

| Method | Params(M) | FLOPs(G) | VoxCeleb1-O EER(%)/MinDCF | VoxCeleb1-E EER(%)/MinDCF | VoxCeleb1-H EER(%)/MinDCF |
|---|---|---|---|---|---|
| CAM++ | 7.18 | 1.73 | 0.88/0.0935 | 0.97/0.1183 | 1.89/0.1971 |
| +DSM | 7.36 | 1.36 | 0.83/0.1091 | 1.01/0.1184 | 1.94/0.2044 |
| ++MFA | 8.15 | 1.37 | 0.83/0.0891 | 0.95/0.1114 | 1.86/0.1922 |

### 3.3. Results analysis

Table 1 shows the experimental results of the proposed and baseline models. For fair comparison, all baseline models involved in the experiment are retrained using the same experimental setups. By comparing with ECAPA-TDNN and ResNet34, it can be found that the proposed LightCAM has achieved slight performance improvements in EER and MinDCF on the O, E, and H test sets of VoxCeleb1. Meanwhile, LightCAM has smaller number of parameters comparing with ECAPA-TDNN. This indicates that, with the goal of model lightweighting, LightCAM retains a slight advantage over mainstream speaker verification systems in terms of performance at a small parameter cost.

Compared to CAM++, LightCAM achieves a further 5.7% improvement in EER for VoxCeleb1-O at the cost of 13.5% of the parameter set, demonstrating the effectiveness of the proposed DSM and MFA methods in improving model performance.

Table.2. Complexity comparison of the model

| Model | Params(M) | FLOPs(G) | RTF |
|---|---|---|---|
| ECAPA-TDNN | 14.65 | 3.96 | 0.042 |
| ResNet34 | 6.63 | 6.84 | 0.057 |
| CAM++ | 7.18 | 1.73 | 0.026 |
| **LightCAM** | **8.15** | **1.37** | **0.017** |

### 3.4. Computational complexity analysis

In this section, we compare the complexity of LightCAM with mainstream ECAPA-TDNN, ResNet34 and CAM++ models in terms of number of parameters, FLOPs and RTF. RTF is evaluated on the CPU device under single-thread condition. Results in Table 2 shows that LightCAM has 55.6% parameters and 34.6% FLOPs compared to ECAPA-TDNN. LightCAM has slightly more parameters compared to ResNet34 but significant fewer FLOPs and RTF. Moreover, LightCAM also reduces FLOPs by 20.8% and RTF by 34.6% compared to CAM++. It is worth noting that LightCAM achieves the fastest inference speed among all mainstream methods. The computational complexity of the proposed model has been greatly improved, making it more suitable for practical applications.

### 3.5. Ablation study

Ablation studies are conducted to explore the specific effects of the proposed improvement method on model performance, parameter quantity, and computational complexity. The experiment used the vanilla CAM++ as the baseline model and applied the proposed DSM and MFA methods. The results are shown in Table 3.

From experimental results, it can be seen that the introduction of DSM, as a method to reduce complexity of the network, has decreased inference speed by 21.4% at the cost of slightly higher EER and MinDCF. MFA is further applied to the benchmark model, then both EER and MinDCF decreased, which are lower than the original CAM++ model, while the inference speed is maintained at the same optimal level. Using these two proposed improvement methods has resulted in both lightweight and improved performance of LightCAM.

### 4. CONCLUSION

Through testing and comparison on benchmark datasets such as VoxCeleb, it can be concluded that LightCAM is superior to other mainstream speaker verification system in the aspect of accuracy and computational complexity. Particularly, it has achieved an obvious reduction in computational complexity while the accuracy in EER and MinDCF has been slightly improved. The experimental data also indicates that the techniques of depthwise separable convolution and multi-scale feature fusion have indeed lower the complexity of the system while enhance the performance. In future work, we will test the performance of the network on more speaker verification datasets and improve the generalization of the model, thereby increasing the possibility of its application in practical scenarios.